\documentclass[journal]{IEEEtran} %[journal] instead of draft

\usepackage{amsmath}

\usepackage[dvipsnames]{xcolor}

\usepackage{float}

\usepackage{graphicx} 
\graphicspath{{images/}}
\DeclareGraphicsExtensions{.pdf,.jpeg,.png}

\usepackage[caption=false]{subfig}

\usepackage{multicol}
\usepackage{multirow}

\usepackage{url}

\usepackage{comment}
\usepackage{verbatim}

\usepackage{booktabs}

\usepackage[font={small,it}]{caption} 
\usepackage{array}

\newcolumntype{L}[1]{>{\raggedright\arraybackslash}p{#1}}

\usepackage{hyperref}
\usepackage{makecell}
\usepackage[nolist]{acronym}

\usepackage[usestackEOL]{stackengine}

\usepackage{CJKutf8}

\usepackage{footmisc}
\usepackage{threeparttable}

\newcommand{\af}[1]{{\color{black}#1}}
\newcommand{\ls}[1]{{\color{black}#1}}
\newcommand{\lsn}[1]{{\color{orange}#1}}
\newcommand{\vp}[1]{{\color{black}#1}}
\newcommand{\pg}[1]{{\color{black}#1}}

\newcommand{\white}[1]{{\color{white}#1}}

\begin{document}

% Paper title
\title{A Comprehensive Study of ImageNet Pre-Training for Historical Document Image Analysis}

%\begin{comment}
\author{
    \IEEEauthorblockN{
        \textbf{Linda~Studer}\IEEEauthorrefmark{1}\IEEEauthorrefmark{2}, \and
        \textbf{Michele~Alberti}\IEEEauthorrefmark{1}\IEEEauthorrefmark{2}, \and
        \textbf{Vinaychandran~Pondenkandath}\IEEEauthorrefmark{1}\IEEEauthorrefmark{2}, \and
        \textbf{Pinar~Goktepe}\IEEEauthorrefmark{1}\IEEEauthorrefmark{2}, \and \\
        \textbf{Thomas~Kolonko}\IEEEauthorrefmark{1}\IEEEauthorrefmark{2}, \and
        \thanks{\IEEEauthorrefmark{1} These authors contributed equally to this work.}
        \textbf{Andreas~Fischer}\IEEEauthorrefmark{2}\IEEEauthorrefmark{3}, \and
        \textbf{Marcus~Liwicki}\IEEEauthorrefmark{2}\IEEEauthorrefmark{4}, \and
        \textbf{Rolf~Ingold}\IEEEauthorrefmark{2}
    }\\
    \vspace{0.2cm}
    \IEEEauthorblockA{
        \IEEEauthorrefmark{2}%
        \textit{Document Image and Voice Analysis Group (DIVA)} \\
        University of Fribourg, Switzerland\\
        \{firstname\}.\{lastname\}@unifr.ch \\
        \vspace{0.15cm}
        \IEEEauthorrefmark{3}%
        \textit{Institute of Complex Systems (iCoSys)} \\
        University of Applied Sciences and Arts Western Switzerland\\
        andreas.fischer@hefr.ch \\
        \vspace{0.15cm}
        \IEEEauthorrefmark{4}%
        \textit{Machine Learning Group} \\
        Lule{\aa} University of Technology, Sweden\\
        marcus.liwicki@ltu.se\\
    }
}
%\end{comment}

\begin{comment}
\author{
    \white{
    \IEEEauthorblockN{
        \textbf{Author A}, \and
        \textbf{Author B}, \and
        \textbf{Author C}, \and
        \textbf{Author D}, \and
        \textbf{Author E}, \and
        \textbf{Author F}, \and
        \textbf{Author G}, \and
        \textbf{Author H}
    }\\
    \vspace{0.2cm}
    \IEEEauthorblockA{
        \IEEEauthorrefmark{2}%
        \textit{affiliation} \\
        affiliation address \\
        affiliation emails \\
        \vspace{0.15cm}
        \IEEEauthorrefmark{3}%
        \textit{affiliation} \\
        affiliation address \\
        affiliation emails \\
        \vspace{0.15cm}
        \IEEEauthorrefmark{4}%
        \textit{affiliation} \\
        affiliation address \\
        affiliation emails \\
    }
    }
}
\end{comment}

% after the title page when using the two side option.
\markboth{}{First author et al. : Title}

% Make the title area
\maketitle

\thispagestyle{empty}

%%*************************************************************************

\begin{acronym}[Bash]
    % General
    \acro{AI}{Artificial Intelligence}
    \acro{AGI}{Artificial General Intelligence}
    \acro{MLP}{Multilayer Perceptron}
    \acro{ANN}{Artificial Neural Network}
    \acro{CNN}{Convolutional Neural Network}
    \acro{KMNIST}{Kuzushiji-MNIST}
    \acro{CLaMM}{Classification of Medieval Handwritings in Latin Scripts}
    \acro{Historical-WI}{Historical Writer Identification}
    \acro{ICDAR}{International Conference on Document Analysis and Recognition}
    \acro{ASPP}{Atrous Spatial Pyramid Pooling}
\end{acronym}

% Abstract
\begin{abstract}

Automatic analysis of scanned historical documents comprises a wide range of image analysis tasks, which are often challenging for machine learning due to a lack of human-annotated learning samples.
With the advent of deep neural networks, a promising way to cope with the lack of training data is to pre-train models on images from a different domain and then fine-tune them on historical documents.
In the current research, a typical example of such cross-domain transfer learning is the use of neural networks that have been pre-trained on the ImageNet database for object recognition.
It remains a mostly open question whether or not this pre-training \af{helps to analyse} historical documents, which have fundamentally different image properties when compared with ImageNet.
In this paper, we present a comprehensive empirical survey on the effect of ImageNet pre-training for diverse historical document analysis tasks, including character recognition, style classification, manuscript dating, semantic segmentation, and content-based retrieval.
While we obtain mixed results for semantic segmentation at pixel-level, we observe a clear trend across different network architectures that ImageNet pre-training has a positive effect on classification as well as content-based retrieval.

\end{abstract}

% Introduction 
\section{Introduction}
\label{toc:introduction}

\begin{table}[!t]
\setlength{\tabcolsep}{2.25pt} % Default value: 6pt
\renewcommand{\arraystretch}{0.6}
\centering
\caption{
This table gives an overview of the different tasks.
The \acf{KMNIST} dataset contains different Hiragana (cursive Japanese) characters, here depicted is \af{the character for ``o''}.
The expected output is the \af{character} label of the image.
The \acf{CLaMM} dataset is annotated for style classification and manuscript dating.  The expected output is the \af{style or date} label for a given image.
The Historical Manuscript Database DIVA-HisDB is annotated for semantic segmentation at pixel level.
The example shown here is from manuscript CB55.
\ls{The output is a \af{segmentation label for each pixel, here shown with different colours}.}
The \acf{Historical-WI} dataset consists of images from different writers. 
Here a section from one of the pages from writer 100 is depicted. 
The output of the network is a ranking of the most similar writers, based on the input image. 
}
%\vspace{-2mm}
\begin{tabular}{cccc}
    \toprule
    Dataset & Input & Task & Output \\ 
    \midrule
    % Classification
    KMNIST & \raisebox{-.5\height}{\includegraphics[width=0.27\columnwidth]{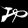}} & \shortstack{Hiragana\\Classification} & \shortstack{\begin{CJK}{UTF8}{min}お\end{CJK} (o)}
    \\ \\ %\midrule
    CLaMM & \raisebox{-.35\height}{\includegraphics[width=0.27\columnwidth]{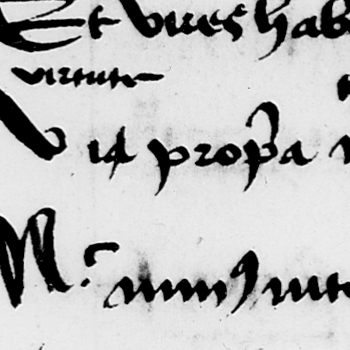}} & \shortstack{Style Classification \\ \\ \\ Manuscript Dating}
    & \shortstack{Semihybrida \\ \\ \\ 1451-1475 C.E.}
    \\ \\ %\midrule
    % Segmentation
    DIVA-HisDB & \raisebox{-.4\height}{\includegraphics[width=0.27\columnwidth]{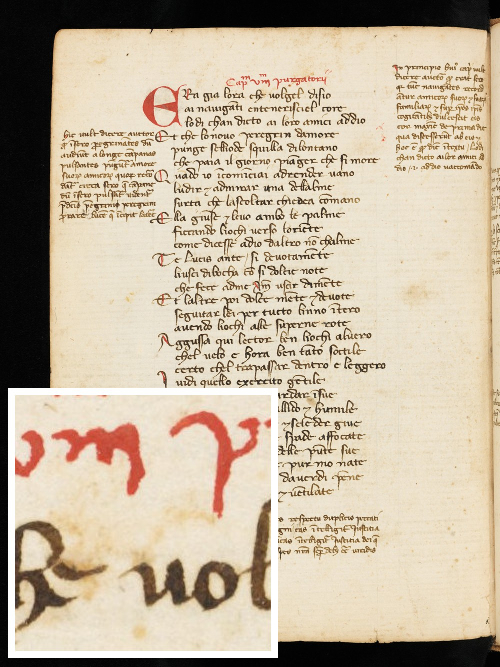}} & \shortstack{Semantic\\Segmentation\\at Pixel Level} & \raisebox{-.4\height}{\includegraphics[width=0.27\columnwidth]{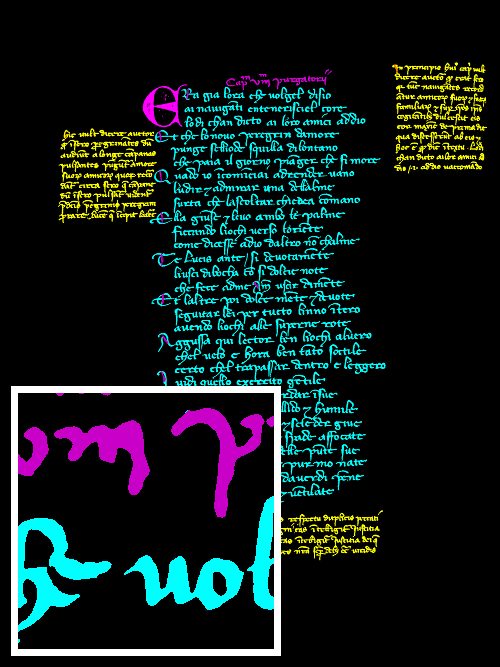}}
    \\ \\ %\midrule
    % Similarity
    Historical WI & \raisebox{-.5\height}{\includegraphics[width=0.27\columnwidth]{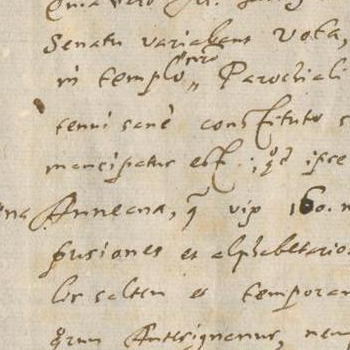}} & \shortstack{Writer Identification} & \shortstack{Writer 100}
    \\
    \bottomrule
\end{tabular}
%\vspace{-3mm}
\label{tab:tasks}
\end{table}
\setlength{\tabcolsep}{6pt}

%\subsection*{Motivation}

Historical documents span centuries of different writing supports (including stone, palm leaf, papyrus, parchment, and paper in different states of decay), writing instruments, languages, scripts, fonts, ornaments, illustrations, and so on. Furthermore, the image acquisition methods may vary substantially depending on the type of document. When performing automatic image analysis for a specific type of document using machine learning, one of the main challenges is to collect a sufficient amount of representative learning samples. In the case of ancient languages and scripts, such annotations often can only be provided by experts in the respective field and are thus costly to obtain.

In recent years, the use of deep neural networks has strongly influenced \af{the} state of the art for historical document analysis. However, deep neural network models have millions of parameters to fine-tune and a random initialization~\cite{glorot2010understanding} may not be the best option when facing a lack of annotated training data. 
Several promising alternatives have been suggested including layer-wise pre-training~\cite{hinton2006fast,alberti2017lda} and transfer learning~\cite{caruana1997multitask}. 
The latter is the main focus of the present paper. 
Transfer learning aims to fine-tune network parameters with respect to another image analysis task -- which features a large amount of annotated training data -- and then use these parameters as an initialisation for the image analysis task at hand.

% Transfer Learning
Transfer learning is a widespread technique in computer vision~\cite{pan2010survey,thrun2012learning}.
Since the publication of large datasets such as ImageNet~\cite{imagenet_cvpr09}, %Cityscapes~\cite{DBLP:journals/corr/CordtsORREBFRS16, 
CIFAR-10 \cite{krizhevsky2014cifar},
PASCAL~\cite{Everingham15}, and COCO~\cite{lin2014microsoft}, many architectures have been trained on them and their weights made publicly available to be used for transfer learning.
Although transfer learning has been around for the last two decades \cite{machinelearning}, it has only become popular in the last years with the breakthrough of \acp{CNN} architectures consistently winning the Large Scale Visual Recognition Challenge (ILSVRC)~\cite{imagenet_cvpr09} competition since 2012~\cite{krizhevsky2012imagenet}. 
% ImageNet is Cool
Pre-training on ImageNet and successive fine-tuning on another dataset has become a widely used practice~\cite{huh2016makes,zeiler2014visualizing}.
It is generally believed that this approach helps to learn good and general features. 

%*************************************************************************************************
\subsection*{Contribution}
% In the end, what do you bring new? Why would one care?
Previous studies mostly focus on fine-tuning on datasets similar to the dataset used for pre-training. Moreover, they only explore a small set of tasks and neural network architectures.
Historical documents have \af{very different image properties when} compared to the natural images found in the ImageNet dataset. It is therefore not immediately intuitive that pre-training a model on ImageNet for historical document analysis will have the same benefits.
%\vspace{10pt}
%\newpage

%In this paper, we investigate the usefulness of ImageNet pre-training in the context of historical document analysis. We choose a variety of different applications, datasets and architectures to provide a comprehensive study of the impact of transfer learning from ImageNet pre-trained models to historical document analysis.
\af{In this paper, we provide a comprehensive empirical study of the impact of transfer learning from ImageNet pre-trained models to historical document analysis. A variety of different applications, datasets and network architectures are taken into account.}
The applications can be grouped into three categories: classification, semantic segmentation \ls{at pixel level} \af{and content-based image retrieval.} 
Most of the datasets we use were published as part of previous \af{competitions at the} \ac{ICDAR}.
\af{Note that the main aim of our study is to investigate the effect of pre-training on relevant and high-quality datasets, and not to outperform the winners of the competitions by optimizing task-specific pre- and post-processing methods.}

% Related Work 
\section{Related Work}
\label{toc:related_work}

% ImageNet is Cool
ImageNet is the most widely used dataset for pre-training and transfer learning.
Popular beliefs as to why ImageNet is particularly suited for this task are its large size, the high number of distinct classes and the close similarity of many of the classes, e.g. \ls{a number of} different dog breeds.

Huh et al. \cite{huh2016makes} examined the impact of various aspects \ls{of ImageNet pre-training and successive fine-tuning on the PASCAL~\cite{Everingham15} dataset, such as dataset size, number of classes, using fine-grained versus coarser class labels and the ratio of images per class.} 
Additionally, they showed that the aforementioned commonly held beliefs are not accurate, and that transfer learning still works well with restrictions, such as only using half of the dataset.

He et al. \cite{he2018rethinking} showed that pre-training on ImageNet speeds up convergence early in training, but that training from scratch will eventually catch up and sometimes even surpass the pre-trained and fine-tuned accuracy.
They further argue that ImageNet pre-training does not automatically give better regularisation and that it shows no benefit when the target tasks or metrics are more sensitive to spatially localised predictions. 
Training from scratch requires different normalisation and regularisation methods as compared to transfer learning.
This can skew results, benefiting the pre-trained paradigm over learning from scratch. 

Similarly, Kornblith et al. \cite{kornblith2018better} showed that although ImageNet pre-training accelerates convergence, it does not necessarily lead to a better performance if run long enough. 
They also investigated how transfer learning relates to the architecture used in the context of image classification.
Their findings suggest that the accuracy increase from ImageNet pre-training fades quickly as the size of the dataset for the task at hand grows larger.  
They conclude that pre-training is and will remain an essential tool in the near future but also highlight clearly that it has limitations. 

% Cross-Domain Transfer Learning
When it comes to cross-domain transfer learning, its usefulness - especially from natural images such as ImageNet - is subject of an open discussion.
There are cases where transfer learning across domains has been proven to be successful \cite{singh2017transforming, pondenkandath2018watermarks}.
In contrast, there is literature suggesting that this technique is harmful to the final performance of the networks.
Tensmeyer et al. \cite{tensmeyer2017convolutional} specifically question the usefulness of transfer learning from ImageNet (natural images of 3D objects) on document analysis, which are 2D entities. 
They argue that feature mappings for natural images fundamentally differ from feature mappings for documents.
They also question whether architectures optimized for natural image classification are a good fit for historical document analysis.
%\ls{However, they do not provide conclusive results and focus only on the AlexNet \cite{krizhevsky2012imagenet} architecture.}
In their work, they do not provide conclusive results, focus only on the AlexNet \cite{krizhevsky2012imagenet} architecture and lack a thorough comparative study.
\vp{The impact of transfer learning from ImageNet was also a topic of discussion at \ac{ICDAR} 2017. The organizers of the Competition on Historical Document Writer Identification~\cite{fiel2017icdar2017} speculated that the competition participant who used a deep learning-based method, may have performed relatively poorly due to their network being initialized with ImageNet pre-trained weights.}

Therefore, a conclusive answer on what the real reasons behind the benefit of transfer learning are and whether these can be harnessed in a cross-domain scenario is yet to be found.

% Experimental Setup
\section{Study Design}%Experimental Setting
\label{toc:experimental_setting}

In this section, we present the details of our \af{empirical study}, namely the tasks we choose to perform, the datasets and network architectures we use and finally the training procedure for each specific task. 
%experimental setting

%***********************************************************************************
%***********************************************************************************
%***********************************************************************************
\subsection{Tasks}
% Explain what are you trying to do. Assume your audience is smart people from the field, so this section might be small.
In this work, we choose three tasks as representatives of \af{some of} the most common challenges in the \af{field of historical document analysis}.
Specifically, our use cases include image classification, \af{semantic segmentation at pixel level and content-based image retrieval}.
We believe that their radically different natures will give a robust estimation of the generality of our findings.
Table \ref{tab:tasks} gives an overview of the input and output of the different tasks.

%***********************************************************************************
\subsubsection{Classification}

This task is well known in the computer vision community and consists of producing one or more descriptive labels for a given input image.
In the context of historical image document analysis this task can be, for example, formulated as character recognition ~\cite{clanuwat2018deep,vamvakas2008complete}, style/script classification ~\cite{al2011recognition,cloppet2017icdar2017} or \af{manuscript dating ~\cite{cloppet2017icdar2017,wahlberg2016historical}}.
We train the networks to minimize the cross-entropy loss function shown below:
\begin{equation}
L(\vec{x}, y) = - log \left( \frac{e^{\vec{x}_y}}{\sum_{i=0}^{n}e^{\vec{x}_i}} \right)
\end{equation}
where $n$ is the number of classes, $x$ is a vector of size $n$ representing the output of the network, and $y = \{1..n\}$ \af{is a scalar} representing the class label, e.g. style of the document.

%***********************************************************************************
\subsubsection{Semantic Segmentation at Pixel Level}

Semantic segmentation at pixel level is a specific case of a classification task.
In this task, each pixel of an input image has to be assigned a label.
This is often performed to analyse the layout of historical documents ~\cite{antonacopoulos2009icdar,antonacopoulos2013icdar}, or as a form of pre-processing for further tasks, e.g. line segmentation \cite{albertivoegtlin2019line}. 
Neural networks for semantic segmentation are trained similarly to the ones for classification, but the architectures employed are different (see section \ref{exp:architecture}).

%***********************************************************************************
\subsubsection{Content-based Image Retrieval}

Image similarity (or content-based image retrieval) is another typical scenario found in computer vision.
In the context of historical document image analysis, this can be seen in the form of writer identification \cite{fiel2017icdar2017}, signature verification \cite{maergner2018} or watermark recognition \cite{pondenkandath2018watermarks}. 
To train the networks for this task, we use the triplet loss approach~\cite{hoffer2015,balntas2016}.
The triplet loss operates on a tuple of three images $\{a, p, n\}$ where $a$ is the anchor (reference), $p$ is the positive sample (an image of the same class as the reference), and $n$ is the negative sample (an image of another class).
The loss function is then defined as:

\begin{equation}
L(\delta_+,\delta_-) = max(\delta_{+} - \delta_{-} + \mu, 0)
\end{equation}

\noindent\ls{where $\delta_+$ and $\delta_-$ are the Euclidean distances between the anchor-positive and anchor-negative pairs in the high dimensional output space of the network and $\mu$ is the margin used.}

%***********************************************************************************
%***********************************************************************************
%***********************************************************************************
\subsection{Datasets}
\label{exp:datsets}
% Explain which dataset you used. Do not hesitate to give some details such as how many classes, #samples and the size of the images (if you use images, that is). DO NOT forget to cite the original source paper for the dataset!

The datasets are available for download through DeepDIVA\footnote{\url{https://bit.ly/DeepDIVA}}.
The image input size depends on the architecture and is described in section \ref{exp:training}. Table \ref{tab:tasks} shows an example image for each dataset.

%***********************************************************************************
\subsubsection{Kuzushiji-MNIST}
The \ac{KMNIST} dataset \cite{clanuwat2018deep} contains grayscale images of ten different Hiragama characters written in \textit{Kuzushiji} (cursive Japanese). 
It is a curated subset of the full Kuzushiji dataset, which was created during the digitisation of around 300'000 old Japanese books. 
The images in \ac{KMNIST} are from 35 classic books printed in the 18th century. The training set contains 7'000 and the test set 1'000 images per class, each of size $28\times28$ pixels.

%***********************************************************************************
\subsubsection{ICDAR2017 Classification of Medieval Handwritings in Latin Scripts}
This dataset was published for the \acf{CLaMM} competition \cite{cloppet2017icdar2017} at the \ac{ICDAR} 2017 conference and includes 3'540 images annotated for style classification and manuscript dating. 
The test set contains 2'000 images. 
The dataset is divided into 12 classes for style classification 
(Caroline, Cursiva, Half-Uncial, Humanistic, Humanistic Cursive, Hybrida, Praegothica, Semihybrida, Semitextualis, Southern Textualis, Textualis, Uncial). 
Each of the 15 classes provided for manuscript dating corresponds to a specific time interval, ranging from 500 C.E. to 1600 C.E.
The competition provided two variations of the dataset, we use the subset that contains images of mixed resolutions and colour representations.

%***********************************************************************************
\subsubsection{ICDAR2017 Competition on Layout Analysis for Challenging Medieval Manuscripts}
The DIVA-HisDB dataset \cite{simistira2016diva} consists of three different medieval manuscripts (CB55, CSG18, CSG863), each containing 50 pixel-wise annotated pages with a size of approximately $4k\times5.5k$ pixels. 
The manuscripts have a challenging layout with four different classes (main text body, decoration, comment and background). 
There is also an additional label for boundary pixels. 
These pixels originate from the labelling process and are background pixels along the border of the text, which are labelled as text. For our purposes, we relabelled these pixels as background. The same training and test dataset split is used as in the \ac{ICDAR} 2017 Competition on Layout Analysis for Challenging Medieval Manuscripts \cite{simistira2017icdar2017}.

%***********************************************************************************
\subsubsection{ICDAR2017 Historical Writer Identification}
The \ac{Historical-WI} dataset \cite{fiel2017icdar2017} consists of colored and binarized versions of handwritten historical documents. 
The training dataset consists of 394 writers with three pages per writer, which gives a total of 1'182 pages. 
The dataset for the competition contains 720 writers and five pages per writer, which makes 3'600 instances in total. 
We use the coloured version of the dataset for our experiments.

%***********************************************************************************
%***********************************************************************************
%***********************************************************************************
\subsection{Model Architectures}
\label{exp:architecture}
% Explain which architecture you have. DO NOT put weird tables with number of layers and what-not details that NOBODY cares about (unless your paper is about a novel architecture of course...)!!! 
% Here you would say something along the lines of "hey we use a model similar to ResNet50 and you can find it on the DeepDIVA repo at \url{link to the point}" and maybe motivate why you chose this model over all possible models. Also: DO NOT forget to cite hte paper of the architecture! (e.g. the ResNet one or so)

For the \af{classification and content-based image retrieval} experiments, we investigate four well-known architectures: VGG19 with batch normalisation (VGG19 BN), Inception V3, ResNet152 and DenseNet121.
A simple architecture with \af{only three} layers is also trained for each task to give a baseline for the performance.
VGG19 uses batch normalisation \cite{simonyan2014very} and consists of alternating blocks of convolutional and max-pooling layers. 
The Inception architecture \cite{Szegedy_2015_CVPR} introduced Inception blocks, which combine different convolutional filters and layers into one block by concatenation.
A second classification head further back in the architecture is added to counteract the vanishing gradient.
Another way to combat the vanishing gradient was introduced with the ResNet architecture \cite{DBLP:journals/corr/HeZRS15}. 
The layers in these type of networks contain direct, additive connections from one layer to a next one, so-called skip connections. 
We use the ResNet152 architecture. 
The idea of skip connections has also been extended to not only connect one layer to the next one, but to have blocks of densely-connected layers.
Unlike for the skip connections, the output is not added but concatenated.
These dense blocks are alternated with $1\times1$ convolutions and max-pooling layers in order to reduce the number of parameters in the model. 
Here, we use the DenseNet121 \cite{DBLP:journals/corr/HuangLW16a}, which features such dense blocks.

For the segmentation experiments, we use two different architectures: SegNet and DeepLabV3.
SegNet \cite{DBLP:journals/corr/BadrinarayananH15} can use any VGG architecture as an encoder, we use VGG19 BN. 
For each encoder layer, there is a decoder layer which uses the max-pooling indices from the respective encoder layer to perform non-linear up-sampling.
The DeepLabV3 architecture \cite{DBLP:journals/corr/ChenPSA17} uses a ResNet as the encoder, ResNet18 in our case, and an \ac{ASPP} as the decoder.
A simple architecture with five layers is also trained on the dataset to give a baseline for the performance.
 
All the architectures are available in DeepDIVA\footnote{\url{https://bit.ly/2VXDIoo}}.

\begin{comment}
\begin{table}[t!]
    \caption{Overview of all the architectures used in this paper. VGG19, Inception V3, ResNet152 and DenseNet121 have been used for the classification and image-based content retrieval task. These networks expect a certain input size. FC-Densent57 (Tiramisu), SegNet and DeepLabV3 are architectures used for segmentation. These architectures can take various input sizes, we sampled random crops of size 256 from the input images.}
    \label{tab:architectures_overview}
    \vskip 0.15in
    \begin{center}
    \begin{small}
    \begin{sc}
    \begin{tabular}{l rc c }
    \toprule
        & \# parameters && input size \\
        \cmidrule{2-2} \cmidrule{4-4}
        VGG19 BN \cite{simonyan2014very} & $140'105'664$ && $224\times224$ \\
        Inception V3 \cite{Szegedy_2015_CVPR} & $24'703'968$ && $299\times299$ \\
        ResNet152 \cite{DBLP:journals/corr/HeZRS15} & $58'406'080$ && $224\times224$ \\
        DenseNet121 \cite{DBLP:journals/corr/HuangLW16a} & $7'085'056$ && $224\times224$ \\
        FC-Densenet57 \cite{jegou2017one} & $1'376'216$ && $256\times256$ \\
        UNet \cite{ronneberger2015u} & $34'573'713$ && $256\times256$ \\
    \bottomrule
    \end{tabular}
    \end{sc}
    \end{small}
    \end{center}
    \vskip -0.1in
\end{table}
\end{comment}

%***********************************************************************************
%***********************************************************************************
%***********************************************************************************
\subsection{Training Procedure}
\label{exp:training}
% This is a section where you want to make sure to provide enough details to reach two goals:
% First and foremost, you want that the regular reader understands what you are doing. In some cases this is trivial, in others it is not.
% Second, you want an interested reader to have enough details to reproduce your experiments. DeepDIVA helps you a lot on this (see the original publication at ICHFR for details about that).

All experiments are run using the DeepDIVA framework \cite{alberti2018deepdiva}. \ls{The models are trained long enough to reach convergence on the training data.}
Each architecture is trained from scratch as well as with ImageNet \cite{imagenet_cvpr09} pre-training to evaluate the effect of pre-training.
All hyper-parameters can be found in our fork of DeepDIVA\footnote{\url{https://bit.ly/2I8c3dX}}.

\subsubsection{Classification}
The architectures described in section \ref{exp:architecture} are trained with data balancing.
Three different classification tasks are performed. 
For the character recognition task on the \ac{KMNIST} dataset, the models are trained for 35 epochs. \ls{The input images are resized to match the required input size of the respective network.}
On the \ac{CLaMM} dataset, the models are trained for 50 epochs for manuscript dating and style classification.
\ls{10 random sections of the required input size of the respective network are sampled from each input image, evaluated and their output is averaged.}

\subsubsection{Semantic Segmentation at Pixel Level}
The architectures described in section \ref{exp:architecture} are trained for 50 epochs.
Since the images are very large, using the whole image as an input for the network is not feasible.
Instead, a total of $60'000$ sections of size $256\times256$ are randomly sampled per training epoch. 
In the test phase, crops of size $256\times256$ are sampled as a sliding window (with $50\%$ overlap) to segment the full image. 
Both architectures used for this task feature encoders, for which ImageNet pre-trained weights are available. 
For the experiments that use pre-training, we initialise only the encoder using these weights, the weights of the decoder are initialised randomly. 

\subsubsection{Content-based Image Retrieval}
The architectures, as described in section \ref{exp:architecture}, are all trained for ten epochs \vp{with $1.5$ million triplets.} 
\vp{All the networks are designed to embed the input images in a $128$-dimensional space.}
The images are randomly cropped to match the required input size of the respective network. 
During training one random section per page is fed to the network. During the test phase, ten random sections of the input image are evaluated, and their output is averaged.
The triplets are regenerated after every epoch.

% Results
\section{Results}
\label{toc:results}

\begin{table*}[t]
    \caption{Accuracy (\%) on the test set for the different architectures trained for the different classification tasks. Character recognition is performed on the \ac{KMNIST} dataset. The reported accuracy is the average along with the standard deviation from five \af{runs}. \af{Style classification and manuscript dating} are preformed on the \ac{CLaMM} dataset.}
    \label{tbl:classification-test-acc}
    \centering
    \begin{small}
    \begin{sc}
    \resizebox{0.99\textwidth}{!}{%
    \begin{tabular}{rccclccclccc}
    \toprule 
    & \multicolumn{3}{c}{Character Recognition}    && %
      \multicolumn{3}{c}{Style Classification} && %
      \multicolumn{3}{c}{Manuscript Dating} \\
    \cmidrule{2-4} \cmidrule{6-8}  \cmidrule{10-12}
    & scratch & pre-trained & $\Delta$ && %
      scratch & pre-trained & $\Delta$ && %
      scratch & pre-trained & $\Delta$ \\ 
    \midrule
    \ls{3-layer cnn}    & 92.98$\pm$0.22 & n/a                     & -      && 12.4 & n/a  & -              && 11.7  & n/a  & -\\
    vgg19 bn     & 98.17$\pm$0.18 & 98.35$\pm$0.15          & +0.18  && 42.5 & 52.1 & +9.6          && 24.0 & 36.1 & +12.1  \\
    inception v3 & 97.82$\pm$0.11 & 98.51$\pm$0.11          & +0.69  && 46.5 & \textbf{55.5} & +9.0  && 24.8 & 35.4 & +10.6 \\
    resnet152    & 97.27$\pm$0.26 & \textbf{98.69}$\pm$0.10 & +1.42  && 39.1 & 49.3 & +10.2           && 20.6 & \textbf{37.9} & +17.3 \\
    densenet121  & 98.64$\pm$0.06 & 98.56$\pm$0.06          & \ls{-0.08}  && 47.3 & 50.9 & +3.6           && 30.7 & 36.4 & +5.7 \\
    \bottomrule
    \end{tabular}}
    \end{sc}
    \end{small}
\end{table*}

\begin{table*}[t]
    \caption{Mean IU (\%) on the test set of the different architectures trained on the three manuscripts of the DIVA-HisDB dataset. For pre-training, only the encoder is initialized with the pre-trained weights from ImageNet.}
    \label{tbl:segm-meaniu-test}
    \centering
    \begin{small}
    \begin{sc}
    \resizebox{0.99\textwidth}{!}{%
    \begin{tabular}{rccclccclccc}
    \toprule 
    & \multicolumn{3}{c}{Manuscript CB55}    && %
      \multicolumn{3}{c}{Manuscript CSG18} && %
      \multicolumn{3}{c}{Manuscript CSG863} \\
    \cmidrule{2-4} \cmidrule{6-8}  \cmidrule{10-12}
    & scratch & pre-trained & $\Delta$ && %
      scratch & pre-trained & $\Delta$ && %
      scratch & pre-trained & $\Delta$ \\ 
    \midrule
    \ls{5-layer cnn} & 55.7          & n/a  & -     && 56.8 & n/a           & -     && 45.6   & n/a           & -\\
    segnet    & 86.9          & 72.9 & -14.0 && 73.0 & \textbf{75.3} & +2.3  && 81.6   & 61.9        & -19.7 \\
    deeplabv3 & \textbf{92.9} & 91.4 & -1.5  && 69.8                 & 73.1 & +3.3  && 85.5 & \textbf{86.7} & +1.2 \\
    \bottomrule
    \end{tabular}}
    \end{sc}
    \end{small}
\end{table*}

\begin{table}[t]
    \caption{Mean average precision (\%) achieved on the test set by the different architectures trained on the Historical-WI dataset for writer identification.}
    \label{tbl:similarity-test-map}
    %\begin{center}
    \centering
    \begin{small}
    \begin{sc}
    \begin{tabular}{rccc}
    \toprule
    & \multicolumn{3}{c}{Writer Identification} \\
    \cmidrule{2-4} 
    & scratch & pre-trained & $\Delta$  \\
    \midrule
    \ls{3-layer cnn}     & 11.4 & n/a   & -     \\
    vgg19 bn      & 14.6 & \pg{24.0} & +9.4  \\ %\tnote{1}
    inception v3  & 9.1  & 26.1  & +17.0 \\
    resnet152     & 24.7 & 22.1  & -2.6  \\
    densenet121   & 27.2 & \textbf{34.6} & +8.2 \\
    \bottomrule
    \end{tabular}
    \end{sc}
    \end{small}
\end{table}

\af{In the following, results are presented for each of the three chosen tasks individually, i.e. classification, semantic segmentation at pixel level and content-based image retrieval.}

%***********************************************************************************
%***********************************************************************************
%***********************************************************************************
\subsection{Classification}
Table \ref{tbl:classification-test-acc} reports the accuracies achieved by the different architectures trained from scratch and with ImageNet pre-training on the \af{\ac{KMNIST} and \ac{CLaMM} datasets}. \af{In general, the network architectures clearly profit from pre-training.}. \ls{ResNet152 shows the \af{largest} increase in performance while DenseNet121 benefits the least.}

%***********************************************************************************
\subsubsection{Optical Character Recognition}
For this task, we report the mean accuracy along with the standard deviation computed over five runs of each experiment. 
Comparing the mean performance of each model using the t-test, the improvement from trained from scratch to pre-trained is statistically significant for VGG19 BN, InceptionV3 and ResNet152.
\ls{DenseNet121 shows a small decrease of 0.08\% with pre-training.
The ResNet152 benefits the most from pre-training with a delta of 1.42\%, which also makes it the best performing model.}

%***********************************************************************************
\subsubsection{Style Classification}
Pre-training leads to a higher accuracy for all architectures with an average increase of 8.1\%. \ls{VGG19 BN, Inception V3 and ResNet152 profit the most from pre-training with an increase in accuracy of around 9.5\%.}

%***********************************************************************************
\subsubsection{Manuscript Dating}
\ls{Pre-training \af{improves} the performance of all the architectures with an average increase of 11.4\%.
ResNet152 shows the largest increase in accuracy with 17.3\%, which also makes it the best performing model.}

%***********************************************************************************
%***********************************************************************************
%***********************************************************************************

\subsection{Semantic Segmentation at Pixel Level: DIVA-HisDB}
The performance of the segmentation models are evaluated \af{with} the layout analysis tool~\cite{alberti2017open} used in the \ac{ICDAR} 2017 competition~\cite{simistira2017icdar2017}. 
The tool computes the mean Intersection over Union (mean IU) between the predicted results and the ground truth. 
Table \ref{tbl:segm-meaniu-test} shows the mean IU achieved by the models on the test set of the three different manuscripts. 
The results regarding the effect of transfer-learning from ImageNet are mixed for both architectures.
On some manuscripts pre-training on ImageNet increases the performance, but on others, the pre-trained network performs much worse.
\ls{In terms of dataset size, semantic segmentation outnumbers the classification and content-based image retrieval by far, as every pixel is a data point.
Kornblith et al. \cite{kornblith2018better} have found that the impact of ImageNet pre-training on the performance of the model becomes smaller the larger the dataset gets. This could explain why pre-training is not beneficial for this task.}

%***********************************************************************************
%***********************************************************************************
%***********************************************************************************
\subsection{Content-based Image Retrieval: Writer Identification}
Table \ref{tbl:similarity-test-map} reports the mean average precision achieved by the different architectures trained from scratch and with pre-training on the Historical-WI dataset.
Pre-training \af{improves} the performance of all models except ResNet152.

% Conclusion
\section{Conclusion}
\label{toc:conclusion}

\ls{For the classification and content-based image retrieval tasks we find a clear trend that cross-domain transfer learning from ImageNet pre-training leads to an improved performance. 
For semantic segmentation at pixel level we obtain mixed results. In some instances pre-training is beneficial but harmed the performance in others.
We speculate that this could be possibly attributed to the larger amount of training data available for semantic segmentation, as each pixel can be considered an individual data point.}

\af{Overall,} in historical document image analysis, the lack of annotated training data is often one of the most limiting factors for machine learning.
Facing this restriction, ImageNet pre-training can significantly help to improve the performance of deep learning models.

\lsn{}

% Appendices 
\section*{Acknowledgment}
The work presented in this paper has been partially supported by the HisDoc~III project funded by the Swiss National Science Foundation with the grant number $205120$\textunderscore$169618$ and by the Rising Tide foundation with the grant number CCR-18-130.

% Bibliography 
\bibliography{biblio}
\bibliographystyle{IEEEtran}

\end{document}